\documentclass[runningheads]{llncs}
\usepackage{graphicx}
\usepackage{amsmath}
\usepackage{amsfonts}
\usepackage{amssymb}
\usepackage{graphics}
\usepackage{xcolor}
\usepackage{caption}
\usepackage{subcaption}
\usepackage{multirow}
\usepackage{mathtools}
\usepackage{float}
\usepackage[linesnumbered,ruled,noend]{algorithm2e}
\begin{document}
\title{Automatic Generation of Individual Fuzzy Cognitive Maps from Longitudinal Data\thanks{This project is supported by the Department of Computer Science \& Software Engineering at Miami University.}}
\titlerunning{Automatic Generation of Individual Fuzzy Cognitive Maps}
%
\author{Maciej K Wozniak\inst{1}\orcidID{0000-0002-3432-6151} \and Samvel Mkhitaryan\inst{2}\orcidID{0000-0003-0667-0542} \and
Philippe J. Giabbanelli\inst{1}\orcidID{0000-0001-6816-355X} }
\authorrunning{Wozniak et al.}
%
\institute{Department of Computer Science \& Software Engineering, Miami University, Oxford OH 45056, USA \and Department of Health Promotion, CAPHRI, Maastricht University, Maastricht, The Netherlands\\
\email{\{wozniamk,giabbapj\}@miamioh.edu, s.mkhitaryan@maastrichtuniversity.nl}\\
\url{http://www.dachb.com} }
\maketitle              
\begin{abstract} 
Fuzzy Cognitive Maps (FCMs) are computational models that represent how factors (nodes) change over discrete interactions based on causal impacts (weighted directed edges) from other factors. This approach has traditionally been used as an aggregate, similarly to System Dynamics, to depict the functioning of a system. There has been a growing interest in taking this aggregate approach at the individual-level, for example by equipping each agent of an Agent-Based Model with its own FCM to express its behavior. Although frameworks and studies have already taken this approach, an ongoing limitation has been the difficulty of creating as many FCMs as there are individuals. Indeed, current studies have been able to create agents whose traits are different, but whose decision-making modules are often identical, thus limiting the behavioral heterogeneity of the simulated population. In this paper, we address this limitation by using Genetic Algorithms to create one FCM for each agent, thus providing the means to automatically create a virtual population with heterogeneous behaviors. Our algorithm builds on prior work from Stach and colleagues by introducing additional constraints into the process and applying it over longitudinal, individual-level data. A case study from a real-world intervention on nutrition confirms that our approach can generate heterogeneous agents that closely follow the trajectories of their real-world human counterparts. Future works include technical improvements such as lowering the computational time of the approach, or case studies in computational intelligence that use our virtual populations to test new behavior change interventions.

\end{abstract}
\section{Introduction}

A \textit{Fuzzy Cognitive Map} (FCM) is an aggregate computational model consisting of factors (weighted labelled nodes), which interact via causal links (directed weighted edges). This approach has been used across a broad range of domains~\cite{felix2019review}, ranging from medical applications~\cite{amirkhani2017review} to socio-environmental systems~\cite{mourhir2021scoping} and engineering~\cite{bakhtavar2021fuzzy}. Many of these applications are motivated by the need to make decisions in complex systems characterized by high uncertainty, feedback loops, and limited access to the detailed temporal datasets (e.g., delays, rates per unit of time) that would support alternative approaches such as System Dynamics (SD). Historically, a computational study would typically result in \textit{one} FCM that depicts the functioning of a system, or \textit{two} FCMs that serve to compare perspectives (e.g., Western views vs. Aboriginal views) about a same social system~\cite{giles2007integrating}. For example, a study on smart cities can yield one FCM that seeks to summarize all potential factors and every relationship~\cite{firmansyah2019identifying}. A specific city would be portrayed as an instance of this overall map, based on specific node values. Similarly, a study on obesity provides a single FCM explaining how weight gain generally works (e.g., stress leads to overeating), and allowing for individuals to be represented through node values (e.g., level of stress)~\cite{giabbanelli2012fuzzy}. 


A growing interest in generating a multitude of FCMs to represent individuals in an Agent-Based Model (ABM) has resulted in questioning this `one-size-fits-all' approach of creating a single map and spawning individuals only by varying node values~\cite{davis2019intersection,giabbanelli2019cofluences}. Intuitively, individuals do \textit{not only} differ on node values (e.g., some are more stressed than others) but \textit{also} on causal links (e.g., some individuals will over-eat when stressed and some will not). In other words, agents can have different traits and also follow different rules. This was empirically confirmed by studies that created FCMs from individual participants, noting that the existence and strength of the causal edges differed widely~\cite{gray2015structure,lavin2018should}. Despite the realization that these virtual populations may be over-simplifying human behaviors, there has been a lack of tools to automatically generate a large population and heterogeneous population. On the one hand, we can ignore the differences in causal links and focus on quickly creating a large number of agents by drawing their node values from calibrated probability density functions, while respecting relevant correlations between distributions~\cite{giabbanelli2014creating}. On the other hand, an abundance of Machine Learning algorithms can automatically set the weight of causal links based on a dataset, but their focus is to create a single FCM rather than a population~\cite{groumpos2018intelligence,papageorgiou2011learning}. There is thus an ongoing need to address the complexity of human behaviors by accounting for heterogeneous traits and a diversity of rules~\cite{mkhitaryan2020dealing}.

Our paper contributes to addressing this need through the automatic creation of different FCMs, thus generating a population of agents that can behave differently. This overarching contribution is realized through two specific goals:
\begin{itemize}
	\item We \textit{propose a new method}, based on Genetic Algorithms (GAs), to generate a virtual population based on the answers of real participants to commonly administered repeated surveys. Although GAs have previously been used to create FCMs from historical multivariate time series~\cite{poczketa2015learning}, we emphasize that our paper proposes the first method to create different FCMs through individual-level data, rather than a single FCM from aggregate data.
	\item We demonstrate that our method leads to a more diverse population than the `one-size-fits-all' approach by applying it to a \textit{real-world scenario in nutrition}. This application relies on our open-source library~\cite{mkhitaryan2021fcmpy} and all scripts used in this case study are publicly accessible at \url{https://osf.io/z543j/}.
\end{itemize}

The remainder of this paper is structured as follows. In Section 2, we succinctly cover Fuzzy Cognitive Maps and how they can be constructed via Genetic Algorithms, while referring the reader to~\cite{stach2005genetic,poczketa2015learning} for detailed technical introductions. In Section 3.1, we present our proposed approach, named `one-for-each'. We describe it through a formal pseudocode, which is explained line-per-line; its implementation as a Jupyter Notebook is available at \url{https://osf.io/z543j/}. Similarly, we describe an approach that represents the current `one-size-fits-all' paradigm in Section 3.2. The two approaches are experimentally compared on a case study in Section 4. Results are discussed in Section 5 together with the limitations of our project and opportunities for follow-up studies.

\section{Background}
\subsection{Fuzzy Cognitive Maps}
A Fuzzy Cognitive Map (FCM) has two components~\cite{felix2019review,giabbanelli2019cofluences}. First, \textit{structurally}, it is a directed, labeled, weighted graph. Each node has a value from 0 (absence) to 1 (full presence); for example, 0 on stress means no stress whatsoever, whereas 1 signifies as much stress as physically possible. Each edge has a weight from -1 to 1 representing causality. For instance, a link $A \rightarrow B$ with a positive weight (e.g., stress $\xrightarrow{0.5}$ depression) states that $A$ causally increases $B$. Mathematically, the \textbf{w}eights of the edges are stored in an adjacency matrix $W$ and the value of the $n$ nodes are stored in a vector whose initially value (i.e., at $t=0$) is $A$.

Second, the nodes' values are \textit{updated} over iterations by Algorithm 1. Nodes are updated \textit{synchronously} based on (i) their current value, (ii) the value of their incoming neighbors, (iii) the weight of the incident edges, and (iv) a clipping or activation function $f$ such as a sigmoid to keep the result within the desired target range of 0 to 1. The update is applied iteratively (Alg. 1, lines 3--4) until one of two stopping conditions is met. The desirable condition is that stabilization has occurred, as target nodes of interest (i.e., outputs of the system) are changing between two consecutive steps by less than a user-defined threshold (Alg. 1., lines 5--6). However, depending on conditions such as the choice of $f$, the system may oscillate or enter a chaotic attractor. The second condition thus sets a hard limit on the maximum number of iterations (Alg. 1., line 2).

\begin{algorithm}[htb]
\caption{\textsc{SimulateFCM}}
\SetKwInOut{Input}{Input}
\Input{input vector $A$ of $n$ concepts, adjacency matrix $W$, clipping function $f$, max. iterations $t$, output set $O$, threshold $\tau$}
\label{alg:simulatefcm}
\DontPrintSemicolon
		$output^0 \leftarrow A$\; 
		\For{$step \in \{1, \ldots, t\}$}{ 
		\For{$i \in \{1, \ldots, n\}$}{
			$output^{step}_i \leftarrow f(output^{step-1}_{i} + \sum_{j=1}^{n} output^{step-1}_j \times W_{j,i})$\;
			}
			\If{$\forall o \in O, |output^{step}_{o} - output^{step-1}_{o}| < \tau$}{
			return $output$
		}
		}
		
		return $output$
\end{algorithm}

\subsection{Genetic Algorithms for Fuzzy Cognitive Maps}
Historically, FCMs were mostly used as expert systems. A panel of subject-matter experts would be conveyed to identify the concept nodes and/or evaluate their causal strength qualitatively (e.g., `low', `medium'). These qualifiers would then be transformed into numerical weights using fuzzy logic~\cite{pedrycz1994triangular}. To improve the accuracy of an FCM, learning algorithms have emerged as an approach that uses data to either fine-tune some of the experts' weights (e.g., nonlinear Hebbian learning) or set all the weights. The latter was pioneered in 2005 when Stach and colleagues used Genetic Algorithms~\cite{stach2005genetic}. Their method, known as real-coded genetic algorithm (RCGA) ``does not require human intervention and provides high quality models that are generated from historical data''~\cite{stach2010divide}. Variations of this method have been proposed, such as the use of a density parameter to force the algorithm in generating a sparse FCM that more closely resembles the map created by experts~\cite{stach2012learning}. Such sparsity mechanism and the use of evolutionary algorithm remain an active area of research to learn FCMs from data~\cite{wang2021learning}.

Figure 1 shows the main steps of a GA (left) and its application to FCMs in particular via the RCGA (right). The algorithm first creates 100 random matrices. \textit{Fitness} is computed for each weight matrix to quantify the proximity between the \textit{final results} generated through that matrix (i.e., the final node values) and the desired outputs $D$ (Figure 2). That is, the RCGA algorithm seeks results that are similar to the goal upon termination; intermediate results are inconsequential. \textit{Crossover} is then applied, with a probability of $0.9$, via a two steps process: (i) start with the first weight matrix and randomly choose the index of an edge (i.e., a cell in the matrix); (ii) from this cell onward, all weights until the end are swapped with the corresponding cells in the next matrix. The process repeats until all matrices have been visited. For each weight matrix, we then apply \textit{mutations} with a probability of $0.5$, which consists of randomly changing the value of selected edges into the interval [-1, 1]. One of two selection mechanisms is randomly chosen for each iteration: either we randomly choose the same number of edges, or we select fewer edges as iterations go. The fitness function is computed again for the mutated matrices. If any of their fitness values is sufficient ($0.99$), then the matrix with the best fitness is returned. Otherwise, the algorithm continues by selecting weight matrices for the next generation with a probability proportional to their fitness. Given this process, the same matrix can be repeatedly chosen, if its fitness is higher than for other matrices. The algorithm stops after at most $100000$ generations, then returning the matrix with highest fitness. Several of these steps are reused (crossover, mutation, stopping conditions) in our proposed solution, hence they are formalized in Algorithm 2.

\begin{figure}[htb]
\centering
\includegraphics[width=0.7\textwidth]{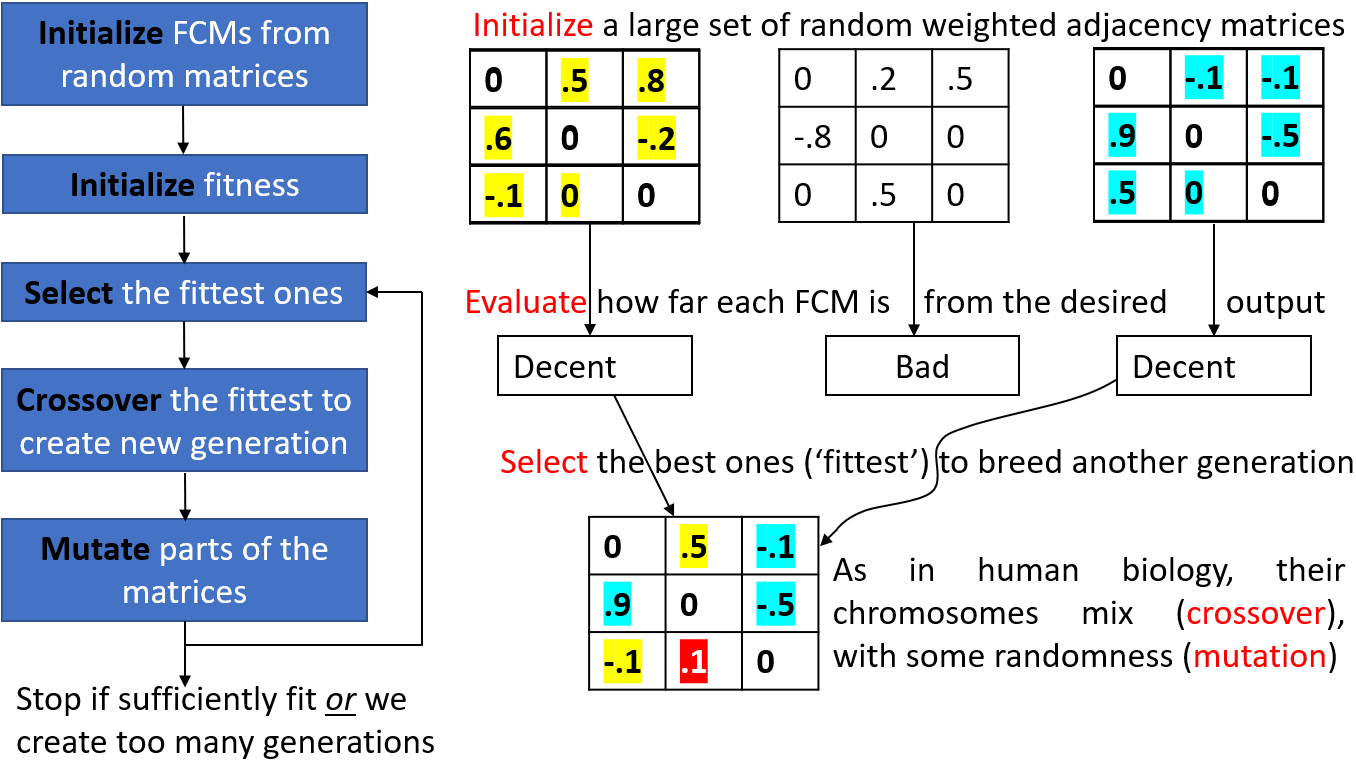}
\caption{\label{fig:cross} Application of a Genetic Algorithm (GA) to Fuzzy Cognitive Maps.}
\end{figure}

\begin{figure}[htb]
\centering
\includegraphics[width=\textwidth]{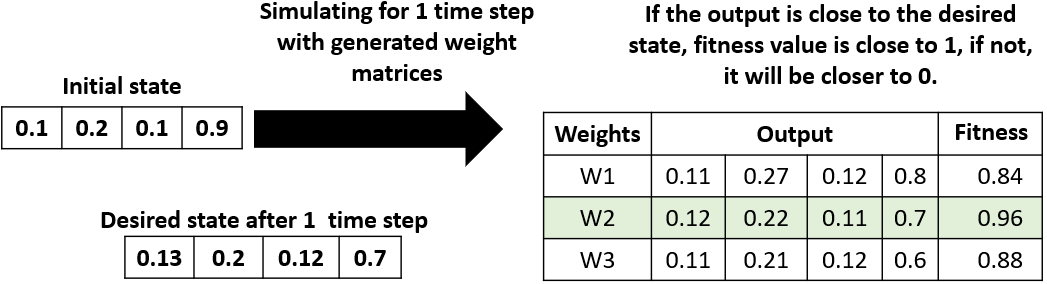}
\caption{\label{fig:rcgafitness} Example to calculate fitness.}
\end{figure}

\section{Methods}
\subsection{Proposed Algorithm to Generate An Individual FCM}
Our `one-for-each' approach is summarized in Algorithm 2. Parameters are listed in Table 1, while noting that they mostly originate from the use of Genetic Algorithms~\cite{stach2005genetic} (Section 2.2); practical choices for parameter values are illustrated via our case study in Section 4. The main steps of Algorithm 2 proceed as follows. The user starts by providing the same parameters (line 1) as in the RCGA (Section 2.2). Then, the initial and desired values of concepts are fed into the algorithm as $t \times n$ 2-D vectors, where \emph{t} is a number of available time steps (three in our case study) and \emph{n} is the number of concepts (lines 2--4). Afterwards, we calculate fitness, with one major difference from the RCGA. Instead of computing fitness only on the final step of an FCM, we compute it at each iteration (lines 5--9). This is essential to ensure that the \textit{trajectory} of an individual is plausible, thus setting more constraints than only replicating the final endpoint. This point is emphasized in Section 4. We check for the two termination conditions (lines 10-14) and, if they are not satisfied, we continue by applying crossover (lines 16--20) and mutations (lines 24--31) before calculating the fitness again (lines 24--31). Weight matrices are selected for the next generation (lines 32--35) and we repeat the process (line 16) until the termination conditions are met. If the fitness of a matrix never reaches the desired $threshold$ within $max\_generations$, we end by returning the matrix with the highest fitness.

\subsection{Algorithms to Generate a Population from Longitudinal Data}
Algorithm 2 creates an FCM that best resembles the \textit{trajectory} of a \textit{single} individual. In order to create a \textit{population}, we thus need to apply that method based on the longitudinal data from \textit{every} individual. This is achieved in Algorithm 3, where we provide the baseline data for each participant ($participant^0$) and up to $T$ consecutive measurements. For example, if participants are seen at 3 months and 6 months, we have $T=2$ as there are two additional measurements. 

\begin{table}[]
\centering
\caption{Parameters of Algorithm 2}
\label{tab:vars}
\scalebox{0.9}{
\begin{tabular}{| p{3cm}|p{2cm}|p{3cm}|p{5cm}|}
\hline
Variable & Notation & Default value & Meaning \\ \hline
Concepts      & $A$          & Vector with values between 0 and 1& Values of concepts (nodes) of FCM         \\ \hline
 Weights       & $W$          & 2-D Array with values between -1 and 1 & Weights of the edges connecting concepts in FCM model         \\ \hline
Number of concepts & $n$ & Integer & Number of concepts (nodes) in FCM model        \\ \hline
Maximum number of generations         & $max\_gen$ & 100000 & Number of generation (iterations) after which RCGA will stop, even if the desired fitness is not reached        \\ \hline
 Maximum fitness        & $threshold$& 0.99 & Minimum fitness value of at least one weight matrix in the generation for the algorithm to stop       \\ \hline
Generation size         & $gen\_size$& 100 & Number of samples (weight matrices) in each generation         \\ \hline
Probability of crossover         & $p\_crossover$& 0.9 & Probability of a pair of samples being chosen for crossover         \\ \hline
Probability of mutation         & $p\_mutation$& 0.5&  Probability of a sample being chosen for mutation\\ \hline
Number of weights chosen for mutation         & $n\_mutation$& 2 &Number of weights chosen for mutation (the same in each time)         \\ \hline
Probability distribution for selection        & $K$& Vector with values between 0 and 1 &Probability distribution created according to fitness values of each sample in the generation          \\ \hline
Uniform distribution        &    $U$& n/a & Uniform distribution of mean 0        \\ \hline
\end{tabular}
}
\end{table}

\scalebox{0.85}{
\begin{algorithm}[H]
\caption{\textsc{MakeIndividualFCM}}
\SetKwInOut{Input}{Input}
\CommentSty{\color{blue}}
\Input{$max\_generations, gen\_size, threshold,$ iterations $t$, number of concepts $n$, desired values of concepts $D$}
\label{alg:rcga}
\DontPrintSemicolon
$\mathbb{W} \leftarrow \{ W^1_{n \times n}, \ldots, W^{gen-size}_{n \times n}\}$ \tcp*{Each matrix is initialized with random weight between -1 and 1}
$\mathbb{A} \leftarrow \{ [0]^1_{t \times n}, \ldots, [0]^{gen\_size}_{t \times n}\}$\;
$fitness \leftarrow [0]_{gen\_size}$\;
\For{$i \in \{1, \ldots, gen\_size\}$}{ 
		$A^i \leftarrow simulateFCM(input\_concepts, W^i, t)$ \tcp*{Simulate using each matrix}
	}
\For{$i \in \{1, \ldots, gen\_size\}$}{
	$error \leftarrow \sum^t_{s=1}\sum^n_{m=1}|A^i_{s,m} - D_{s,m}|$ \tcp*{Calculating fitness for each W}
	$fitness^i \leftarrow \frac{1}{100 \times error + 1}$
}
\If{$max(fitness_{i \in gen\_size}) \geq threshold$}{
	$i \leftarrow 0$ \tcp*{Checking for termination condition}
	\While{$fitness^i < threshold$}{
		$i \leftarrow i+1$
	}
	return $(W^i, fitness^i)$\;
}
\For{$step \in \{1, \ldots, max\_generations\}$}{

	\For{$i \in [1, 3, 5, \ldots, gen\_size]$}{
		\If{p\_crossover}{			
			$idr \leftarrow U(1,n)$ \tcp*{Swapping edges between 2 weight matrices}
			$idc \leftarrow U(1, n )$\;
			$W^i_{\substack{idr \leq i \leq n \\ idc \leq j \leq n}} \leftrightarrow  W^{i+1}_{\substack{idr \leq i \leq n \\ idc \leq j \leq n}}$
		}
		\If{p\_mutation}{
			\For{$i \in \{1, \ldots, n\_mutations\}$}{
				$W^i_{U(1, n ), U(1,n)} \leftarrow U(\{-1,-0.99,\ldots,1\})$\;
			}
		}		
	}
		\For{$i \in \{1, \ldots, gen\_size\}$}{
			$error \leftarrow \sum^t_{s=1}\sum^n_{m=1}|A^i_{s,m} - D_{s,m}|$\;
			$fitness^i \leftarrow \frac{1}{100 \times error + 1}$
		}
		\If{
		$max(fitness_{i \in gen\_size}) \geq threshold$}{
			$i \leftarrow 0$\; \tcp*{Checking for termination condition}
			\While{$fitness^i < threshold$}{
				$i \leftarrow +1$
			}
			return $(W^i, fitness^i)$\;
		}
		$\mathbb{W}\_buffer \leftarrow \mathbb{W}$\;
	\For{$i \in \{1, \ldots, gen\_size\}$}{  
		$idx \leftarrow K(1, gen\_size)$ \tcp*{Probability distribution according to $\frac{fitness}{\sum fitness}$}
		$W\_buffer^i \leftarrow W^{idx}$\;
	}
	$ W \leftarrow W\_buffer$\;
}
	$j \leftarrow 0$ \tcp*{If the $threshold$ value of fitness was not achieved before $max\_generation$ iterations, return W with the highest fitness value}
	$max\_value \leftarrow 0$\;
	\For{$i \in \{1, \ldots, gen\_size\}$}{
		\If{$fitness^i > max\_value$}{
		 $j \leftarrow i$\;
		 $max\_value \leftarrow fitness^i$
		}
	}
	return $(W^j, fitness^j)$\;
\end{algorithm}
}

Although our motivation is to create a virtual population that replicates the heterogeneity of behaviors found in real-world individuals, there are several reasons for which a simplification might be used by taking an `average' individual profile and scaling it to an entire virtual population. For example, generic or `one-size-fits-all' interventions are designed around the notion of an average profile~\cite{epstein1998treatment}. In addition, health data sharing may employ privacy-preserving aggregation algorithms, or only national statistics may be available~\cite{giabbanelli2014modelling}; in either cases, virtual agents may have to be created based on an average archetype. Algorithm 4 captures this situation and serves to empirically establish the difference with our approach; that is, having a `one-for-each' algorithm alongside a `one-size-fits-all' version will let us examine the value-proposition of our approach through a case study. If the difference between the two algorithms is negligible within the application context, then the computational expense of Algorithm 3 would not be warranted. Otherwise, if the difference is noteworthy, then tool will be able to address a practical need.

\begin{algorithm}[htb]
\caption{\textsc{One-For-Each}}
\SetKwInOut{Input}{Input}
\Input{$longitudinalData$, number of participants $p$, number of concepts $n$, follow-up measurements $T$}
\label{alg:oneforeach}
\DontPrintSemicolon
		$Wouts \leftarrow [0]_{p \times n \times n}$\;
		\For{$participant \in longitudinalData$}{
		
		$maxfit \leftarrow 0$\;
		\For{$i \in \{1, \ldots, 100\}$}{
			$W, fitness \leftarrow rcga(1000000, 100, 0.99, participant^{0}, participant^{1,\ldots,T}, T$)\;
				\If{$fitness > maxfit$}{
						$Wouts^i \leftarrow W$\;
				}
			}
		}
		return $Wouts$
\end{algorithm}

\begin{algorithm}[htb]
\caption{\textsc{One-Fits-All}}
\SetKwInOut{Input}{Input}
\Input{$longitudinalData$, number of participants $p$, number of concepts $n$, follow-up measurements $T$}
\label{alg:onefitsall}
\DontPrintSemicolon
		$meanValues \leftarrow \frac{\sum (longitudinalData, axis \leftarrow 1)}{p}$\;
		$maxfit \leftarrow 0$\;
		$Wout \leftarrow \emptyset$\;
		\For{$i \in \{1, \ldots, 100\}$}{
			$W, fitness \leftarrow rcga(1000000, 100, 0.99, meanValues^{T0}, meanValues^{T1,T2}, t$)\;
			\If{$fitness > maxfit$}{
					$Wout \leftarrow W$\;
			}
		}	
		return $Wout$
\end{algorithm}

\section{Case Study}

\subsection{Overview}
Our case study is an intervention on promoting healthy eating among adults in the Netherlands, with a focus on fruit intake~\cite{springvloet2015short}. Sixteen concepts were identified as salient in this application context~\cite{springvloet2015short}, thus we have $n=16$ concepts in our FCMs, with one of them (`daily fruit intake') serving as outcome. Each concept is listed in Table 2, with a unique identifier (left column) used for brevity when plotting results. A total of 1149 individuals took part in the study, out of whom 722 were surveyed two more times (at one month and at four months) beyond the baseline measurement and provided complete data. Consequently, our goal is to generate $722$ FCMs that follow the trajectory of each individual, from the baseline to the two ensuing measurements. 

In Section 4.2, we empirically examine whether our proposed solution (Algorithm 2) is more suited to generate one individual than the previous works (RCGA). Then, in Section 4.3, we assess our ability at generating a heterogeneous population via Algorithm 3, while using Algorithm 4 for comparison. %

\begin{table}[htb]
\centering
\caption{FCM concepts in our case study were chosen by experts in psychology and nutrition to cover relevant domains and subdomains. The unique ID (\#, left column) is used to refer to each concept in figures 4--5.}
\label{tab:samvel}
\begin{tabular}{|p{0.5cm}|p{2cm}p{2cm}|p{7cm}|}
\hline
\textbf{\#}        & \multicolumn{1}{l|}{\textbf{Domain}}                  & \multicolumn{1}{l|}{\textit{\textbf{Sub-domain}}} & \textbf{Question}                                                                                                                                                                           \\ \hline
0                  & \multicolumn{2}{c|}{Perceived Intake/Awareness}                                                           & How many fruits do you think you eat?                                                                                                                                                       \\ \hline
1                  & \multicolumn{1}{c|}{\multirow{2}{*}{Attitude}}        & \textit{Health}                                   & \multirow{2}{*}{\begin{tabular}[c]{@{}l@{}}I think eating two pieces of fruit per day is very...\\ (Unhealthy--Healthy) / (Cheap--Expensive)\end{tabular}}                 \\ \cline{1-1} \cline{3-3}
2                  & \multicolumn{1}{c|}{}                                 & \textit{Cost}                                     &                                                                                                                                                                                             \\ \hline
3                  & \multicolumn{1}{c|}{\multirow{2}{*}{Self-efficacy}}   & \textit{Ability}                                  & Do you think you can eat more fruit per day in the next six months if you really want to?                                                                                                   \\ \cline{1-1} \cline{3-4} 
4                  & \multicolumn{1}{c|}{}                                 & \textit{Difficulty}                               & How difficult or easy do you think it is to eat more fruit in the next six months?                                                                                                          \\ \hline
\multirow{2}{*}{5} & \multicolumn{1}{l|}{\multirow{3}{*}{Social norms}}    & \multicolumn{1}{l|}{\multirow{2}{*}{\textit{Normative}}}   & \multirow{3}{*}{\begin{tabular}[c]{@{}l@{}}Most people who are important to me... \\ (think I should eat two pieces of fruit per day)\\ (consume two pieces of fruit per day)\end{tabular}} \\
                   & \multicolumn{1}{l|}{}                                 & \multicolumn{1}{l|}{}                             &                                                                                                                                                                                             \\ \cline{1-1} \cline{3-3}
6                  & \multicolumn{1}{l|}{}                                 & \textit{Modeling}                                 &                                                                                                                                                                                             \\ \hline
7                  & \multicolumn{2}{c|}{Intention}                                                                            & Do you intend to eat two pieces of fruit per day?                                                                                                                                           \\ \hline
8                  & \multicolumn{1}{c|}{\multirow{3}{*}{Action Planning}} & \textit{When}                                     & \multirow{3}{*}{\begin{tabular}[c]{@{}l@{}}I have a clear plan for... (when I am going to eat\\ more fruit) / (which fruit I am going to eat more\\ or less) / (how many fruit I am going to eat)\end{tabular}}                         \\ \cline{1-1} \cline{3-3}
9                  & \multicolumn{1}{c|}{}                                 & \textit{Which}                                    &                                                                                                                                                                                             \\ \cline{1-1} \cline{3-3}
10                 & \multicolumn{1}{c|}{}                                 & \textit{How many}                                 &                                                                                                                                                                                             \\ \hline
11                  & \multicolumn{1}{c|}{\multirow{4}{*}{Coping Planning}} & \textit{Interferences}                            & \multirow{4}{*}{\begin{tabular}[c]{@{}l@{}}I have a clear plan for what I am going to do...\\ (when something interferes with my plans to eat\\ more fruit) / (in situations in which it is difficult\\ to eat more fruit)\end{tabular}} \\ \cline{1-1} \cline{3-3}
\multirow{3}{*}{12} & \multicolumn{1}{c|}{}                                 & \multirow{3}{*}{\textit{Difficulty}}              &                                                                                                                                                                                                                                          \\
                    & \multicolumn{1}{c|}{}                                 &                                                   &                                                                                                                                                                                                                                          \\
                    & \multicolumn{1}{c|}{}                                 &                                                   &                                                                                                                                                                                                                                          \\ \hline

13                 & \multicolumn{1}{c|}{\multirow{2}{*}{Environment}}     & \textit{Amount}                                   & How often do you have fruit products available at home?                                                                                                                                     \\ \cline{1-1} \cline{3-4} 
14                 & \multicolumn{1}{c|}{}                                 & \textit{Location}                                 & Where do you store the fruit products at home?                                                                                                                                              \\ \hline
15                 & \multicolumn{2}{c|}{Daily fruit intake (pieces)}                                                          & Pieces of fruit per day, including citrus fruit, other fruit and fruit juice                                                                                                                \\ \hline
\end{tabular}
\end{table}

\subsection{Comparing Algorithm 2 with the original RCGA procedure}
One essential difference in our proposed Algorithm 2 compared to the original RCGA procedure is to further constraint the notion of fitness by requiring a fit at \textit{each iteration} of the FCM (to follow each participant) rather than only at the end. This raises several questions. First, is there a difference between the FCMs produced? Second, how closely do they follow the ground truth data? These questions are answered by Figure 3. We see that there is indeed a difference between the FCMs (original in green; ours in orange) across concepts. Most interestingly, we note that the difference is not limited to the intermediate steps, but also carries onto the final results. Our proposed algorithm arrives at an accurate endpoint in the last iteration \textit{and} and remains within 0.01 (on a scale of 0 to 1) of the ground truth value throughout the individual's trajectory. 

\begin{figure}[htb]
\centering
\includegraphics[width=\textwidth]{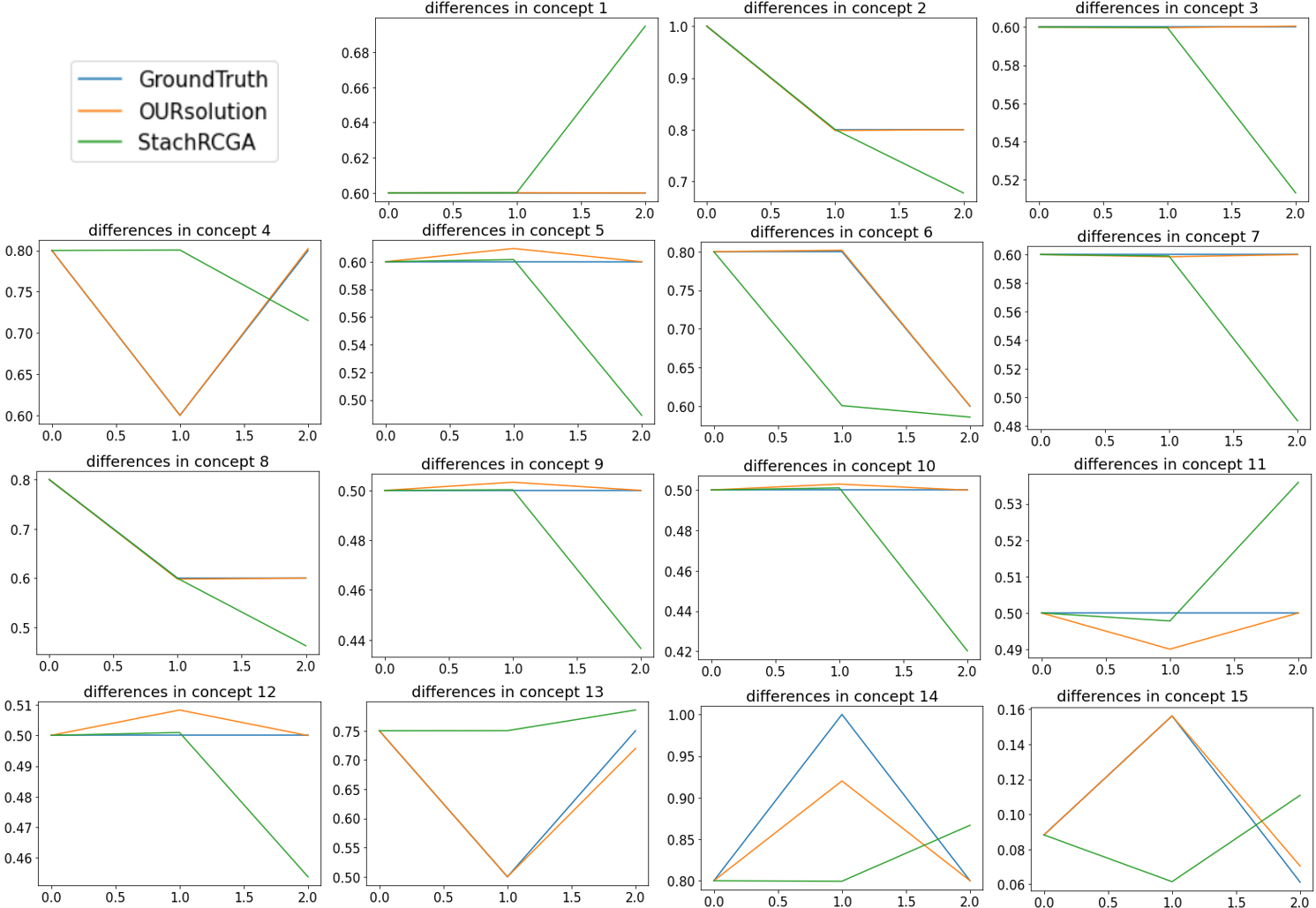}
\caption[Comparing ground truth data (blue) and data generated by original RCGA.]{\label{fig:newrcga} Comparing ground truth data (blue), the original RCGA (green), and our proposed solution (orange) on all concepts of an individual. The two iterations are shown on the x-axis and the concept values on the y-axis. Concept \#0 is not included as it is constant. When the blue line cannot be seen, it overlaps with the orange line (i.e., our solution exactly matches the ground truth).}
\end{figure}

In contrast, the original RCGA procedure is always further from the endpoint \textit{in addition to} being occasionally very far from intermediate steps. For example, on concept \#15 (which is the output of the FCM hence a key concept), the RCGA starts by going down instead of going up, and then goes down instead of up, ending above the desired state as a series of erroneous directions. Wide fluctuations are similarly observed on most other concepts.

\subsection{One-for-each vs. one-size-fits-all}



\begin{figure}[htb]
\centering
\includegraphics[width=\textwidth]{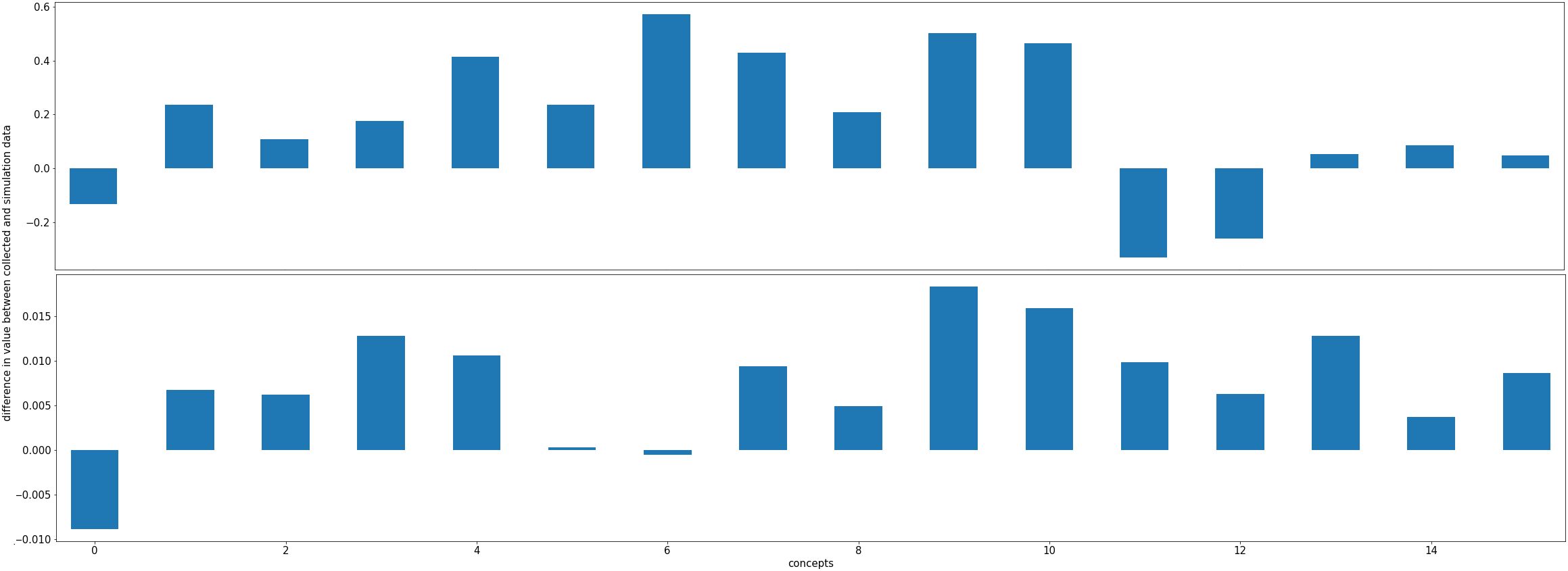}
\caption[Mean difference between ground truth and data generated.]{\label{fig:diffonefitsall} Mean difference between ground truth and data generated using weight matrices obtained by Algorithm 4 (`one-size-fits-all'; top) and Algorithm 3 (`one-for-each'), calculated over 100 randomly selected individuals. Note the \textit{different scales of the y-axis}; top errors range from -0.4 to 0.6 while bottom errors are within -0.01 to 0.02.}
\end{figure}

By applying Algorithm 3, we used the responses of each participant to generate a corresponding FCM, that is, a personalized model that describes their behavior. For analysis, 100 individuals were randomly selected and simulated over two time steps, in line with the empirical data. To assess the quality of this virtual population, we applied Algorithm 4 for comparison. The key difference is that Algorithm 4 is given the \textit{average} participant responses, thus assuming that individuals are distributed around an average behavior. Algorithm 4 was called 100 times, thus leveraging the randomness of the underlying genetic algorithm to generate a diverse population. The two solutions are compared in Figure 5. When using Algorithm 3, the error is within 0.01 for most concepts, with a peak at approximately 0.015 (on action planning). In contrast, using Algorithm 4, the error is an order of magnitude higher, with many concepts being erroneous by a margin of 0.2--0.4 and even a peak at almost 0.6 on social norms. In short, our approach to simulating a heterogeneous population is confirmed, as it yields accurate models. The computational costs of generating individuals are also justified, given that the computationally cheaper approach of generating an 'average' model ends up being wrong for most of the population \textit{in our application context}.

Note that Algorithm 4 \textit{may} be useful in settings where the population is very homogeneous and \textit{not} exposed to an intervention. Indeed, if there are even small differences in baseline \textit{and} behavior change taking place, then individuals may vastly different trajectories~\cite{resnicow2006chaotic}. This notion of `quantum leap for health promotion' is confirmed in our sample. Using the approach by D'Agostino and Pearson's~\cite{pearson1977tests}, we tested the null hypothesis that our participants follow a normal distribution on each concept, at each iteration. At baseline ($t=0$), we found that all concepts but two (\#1, \#15) were normally distributed. However, as participants went through the intervention, \textit{none} of the concepts were normally distributed at later measurements ($t=1$, $t=2$), which justifies the creation of individual models that fit each person's trajectory instead of assuming that an initially normal distribution will remain unchanged.

\section{Discussion}

Fuzzy Cognitive Maps (FCMs) have long been used in fields such as environmental management or medicine~\cite{amirkhani2017review,mourhir2021scoping}, where a single map serves to define how a system functions~\cite{firmansyah2019identifying}. The potential of using this aggregate-level modeling technique to capture how \textit{individuals} function has been researched more recently~\cite{giabbanelli2019cofluences,davis2019intersection}. One key challenge lies in generating enough FCMs to capture the heterogeneity of a population. As FCMs have historically served to synthesize the perspectives of experts, it would be possible to take the viewpoint of each expert and create an FCM accordingly. However, there would only be as many `behavioral archetypes' as there are experts, which still drastically limit the heterogeneity of a population and may not fully match how individuals truly operate. Although Machine Learning (ML) has increasingly been used with simulations~\cite{giabbanelli2019solving} and algorithms exist to train \textit{one} FCM from data~\cite{stach2005genetic,groumpos2018intelligence}, there was no solution to create a large number of FCMs. Our manuscript has sought to address this gap by modifying a Genetic Algorithm~\cite{stach2005genetic,stach2012learning} and applying it on data collected from participants over time. This type of individual-level longitudinal data is commonplace in the study of human behaviors, thus our proposed solution has broad applicability to create virtual populations with precisely calibrated behaviors. Our experimental evaluation on a case study from nutrition confirms that our algorithm can closely follow the trajectories of individuals and even outperforms previous solutions on matching their final state. The case study also illustrated the value of creating a population based on individuals instead of assuming an `average profile', which may hold at baseline but quickly ceases to apply as participants experience a behavior change intervention.

One limitation of this approach is its computational cost. For each individual, 100 matrices are evaluated for their fit, then selected and mutated over up to 100,000 steps. The process is repeated several times to ensure the best fitness. Even at the scale of a small population of 257 individuals (in our case study), a High Performance Computing Cluster was necessary to train the FCMs. An important follow-up study would thus consist of identifying the parameters to which the algorithm is most sensitive (Table 1) and reduce the other ones to lower the total number of combinations computed. In addition to this classic approach to simulation-based optimization, additional constraints can be placed on the algorithm to more quickly guide the generation of relevant FCMs, such as the use of sparsity to discard candidate matrices~\cite{wang2021learning}. From an application standpoint, additional evaluations from other fields of human behavior would complement our case study. In particular, experiments with different trajectories can be an interesting application to assess the ability of our algorithm at closely following variations. 

Ultimately, the main purpose in building a simulation is to \textit{use it}. As we now have a procedure to generate a virtual population with a code-base that is open to the research community, we look forward to case studies in which these populations are used to test new interventions, thus continuing to use simulations to develop insight in human behaviors and find effective policies.

\section{Acknowledgements}
We thank Dr Jens Mueller for his assistance with the RedHawk high-performance computing cluster at Miami University. We also benefited from the feedback of Drs Rik Crutzen, Nanne K. de Vries, and Anke Oenema.
 
\bibliographystyle{splncs04}
\bibliography{bibliography}

\end{document}